\documentclass[runningheads]{llncs}

\usepackage[T1]{fontenc}
\usepackage{graphicx,verbatim}
\usepackage{xcolor}
\usepackage{booktabs}
\usepackage{colortbl}
\usepackage{amsmath}
\usepackage{amssymb}
\usepackage{microtype}
\usepackage{subcaption}
\usepackage{algorithm}
\usepackage{algorithmic}
\usepackage{marvosym}
\usepackage[colorlinks,linkcolor=red, anchorcolor=blue, citecolor=blue, urlcolor=magenta]{hyperref}
\newcommand{\myparagraph}[1]{\vspace{0pt}\noindent{\bf{#1}}~}

\begin{document}

\title{FreeBridge: Variational Schr\"odinger Bridges for Cellular Transition Dynamics}

\author{%
Xurui Wang\inst{1,2}\thanks{Equal contribution.}\and
Qin Ren\inst{1}\protect\footnotemark[1]\and
Jun Ma\inst{3} \and
Haibin Ling\inst{1} \and
Chenyu You\inst{1}\thanks{Corresponding author.}%
}
\titlerunning{FreeBridge: Variational Schr\"odinger Bridges}
\authorrunning{X. Wang, Q. Ren et al.}
\institute{%
Stony Brook University, Stony Brook, NY, USA   \\ 
\email{chenyu.you@stonybrook.edu} \and
University of Toronto, Toronto, ON, Canada \and
University Health Network, Toronto, ON, Canada
}

\maketitle

\begin{abstract}
High-content imaging assays quantify cellular responses to chemical and genetic perturbations, yet continuous trajectories of individual cells are unobservable because cells are chemically fixed at acquisition. Perturbation modeling therefore reduces to inferring stochastic transport between control and treated populations observed only as separate marginals. While recent generative models achieve strong endpoint alignment, boundary consistency does not determine intermediate evolution: multiple stochastic processes may connect identical marginals while traversing regions unsupported by observed single-cell morphologies. We introduce \textbf{FreeBridge}, a Schr\"odinger Bridge formulation for single-cell transition modeling under endpoint-only supervision. FreeBridge defines atomic states as instance-segmented single-cell representations, establishing a fixed cellular manifold, and learns stochastic transport constrained within this geometry via empirical latent support regularization. Across BBBC021, RxRx1, and JUMP, FreeBridge maintains competitive or improved endpoint fidelity and mechanism-of-action retention under a unified evaluation protocol; on BBBC021, it further reduces intermediate support violations. These findings highlight the importance of geometric grounding for biologically interpretable perturbation dynamics. Project page: \url{https://y-research-sbu.github.io/FreeBridge/}.

\end{abstract}


\section{Introduction}
Building a virtual cell capable of simulating cellular responses \emph{in silico} has long been a central aspiration in quantitative cell biology~\cite{slepchenko2003quantitative}.
Image-based profiling enables large-scale measurement of morphological responses under chemical and genetic perturbations~\cite{carpenter2007image,ljosa2012annotated}.
The emergence of large perturbation atlases such as RxRx1 and JUMP~\cite{sypetkowski2023rxrx1,chandrasekaran2023jump}, together with recent efforts to integrate artificial intelligence into virtual cell construction~\cite{bunne2024build}, creates an opportunity to model perturbation dynamics directly from imaging data.

\begin{figure}[t]
\centering
\includegraphics[width=0.98\textwidth]{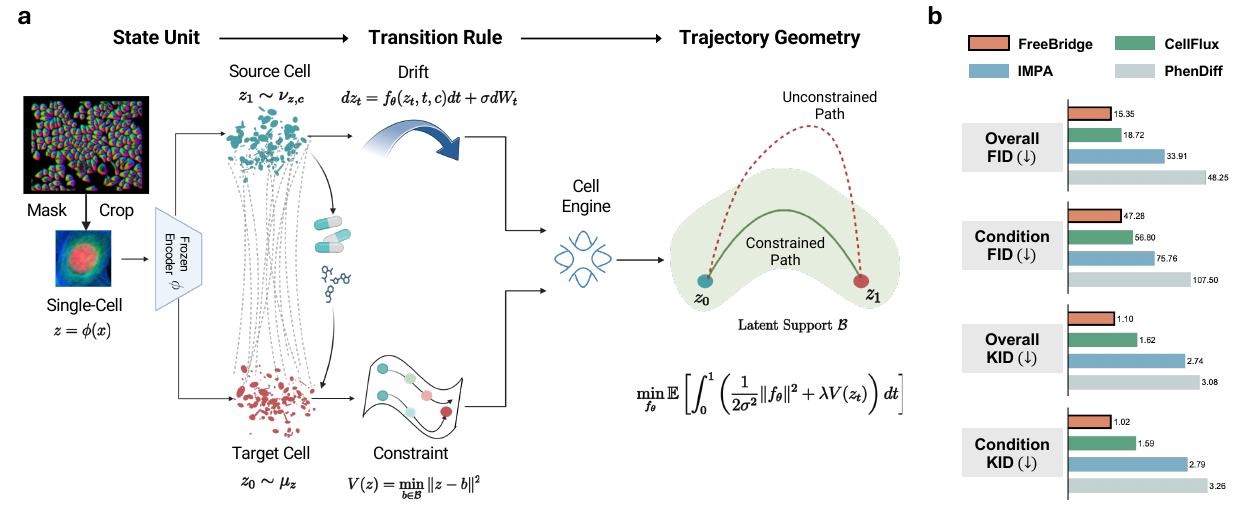}
\vspace{-5pt}
\caption{
\textbf{Overview of FreeBridge.}
(a) Multi-cell microscopy images are segmented into atomic single-cell states, forming the State Unit.
Each crop is embedded as $z=\phi(x)$, defining control and perturbed endpoint distributions.
On this fixed latent geometry, FreeBridge learns a time-conditioned drift $u_\theta(z,t,c)$ to model Schr\"odinger Bridge transport, with an empirical support cost $V(z)$ regularizing intermediate states.
(b) On BBBC021, FreeBridge achieves lower overall and condition-level FID and KID than PhenDiff, IMPA, and CellFlux under the unified single-cell protocol.
}
\vspace{-10pt}
\label{fig:overview}
\end{figure}

Recent generative models learn mappings between control and perturbed distributions.
Diffusion models~\cite{ho2020denoising,song2020score,ren2025scale,sun2025ouroboros}, flow matching~\cite{lipman2022flow,liu2022flow,wang2026let}, and morphology predictors including PhenDiff, IMPA, and CellFlux~\cite{bourou2024phendiff,palma2025predicting,zhang2025cellflux} achieve high endpoint fidelity by framing perturbation prediction as distribution-to-distribution transformation.
However, high-content assays measure chemically fixed cells; individual trajectories are unobservable, and only marginal endpoint distributions are available.
Infinitely many stochastic processes can connect identical marginals~\cite{chen2016relation,leonard2013survey}, so endpoint alignment does not determine intermediate evolution.

Most existing methods optimize distribution-level objectives to match control and perturbed populations~\cite{ho2020denoising,lipman2022flow,bourou2024phendiff,zhang2025cellflux,palma2025predicting}.
While effective for endpoint realism, such formulations leave intermediate cellular structure underdetermined.
The notion of an atomic cellular state is often implicit and entangled with population-level statistics, so intermediate samples may satisfy FID~\cite{heusel2017gans} or KID~\cite{binkowski2018demystifying} while traversing regions weakly supported by observed single-cell morphologies.

A perturbation model therefore requires an explicit definition of admissible cellular states together with a transition rule constrained within that state space.
We describe this structure as a \emph{Cell Engine}, comprising a state unit and a transition mechanism.
Atomic states are instantiated as instance-segmented single-cell representations~\cite{stringer2021cellpose}, establishing a geometrically grounded manifold independent of population composition.
On this manifold, perturbation dynamics are modeled through a Schr\"odinger Bridge-inspired stochastic transport process~\cite{chen2016relation,leonard2013survey}, regularized by empirical latent support constraints grounded in nonparametric density estimation~\cite{silverman1986density,biau2015lectures}.

Based on this formulation, we develop \textbf{FreeBridge}, a constrained stochastic transport framework for single-cell transition modeling under endpoint-only supervision. An overview of FreeBridge is shown in Fig.~\ref{fig:overview}. By explicitly separating state specification from stochastic transport and grounding stochastic evolution within an empirically supported cellular manifold, FreeBridge constrains intermediate trajectories while preserving endpoint fidelity. Evaluations on BBBC021~\cite{ljosa2012annotated}, RxRx1~\cite{sypetkowski2023rxrx1}, and JUMP~\cite{chandrasekaran2023jump} show competitive or improved generative performance relative to prior distribution-alignment approaches~\cite{bourou2024phendiff,palma2025predicting,zhang2025cellflux}; on BBBC021, FreeBridge further reduces intermediate support violations.

\section{Method}
\label{sec:method}

\subsection{Overview: FreeBridge as \textit{Cell Engine}}
High-content screening provides endpoint populations of single cells under control and perturbed conditions. Because imaging requires chemical fixation, per-cell temporal trajectories are not observable. Perturbation modeling is therefore formulated as an endpoint-constrained stochastic transport problem between empirical single-cell distributions rather than as identity-preserving evolution.

As illustrated in Fig.~\ref{fig:overview}(a), we develop \textbf{FreeBridge} under a \emph{Cell Engine} formulation that separates cellular state specification from stochastic transport. The \textbf{State Unit} defines the admissible geometry of valid single-cell representations. The \textbf{Transition Rule} models entropy-regularized stochastic transport within that fixed geometry. This separation ensures that dynamical learning does not implicitly redefine the underlying notion of a biologically valid cellular state.

\subsection{State Unit: Explicit Single-Cell Geometry}
The State Unit specifies the atomic representation of a cellular state. We extract instance-segmented single-cell crops using Cellpose~\cite{stringer2021cellpose}, ensuring that each observation corresponds to a physically identifiable cell rather than a field-level mixture.
Each crop $x$ is embedded using a fixed encoder $\phi : \mathcal X \rightarrow \mathbb R^d$, producing a latent representation: $z = \phi(x)$.
The encoder is frozen during transport learning so that the latent geometry is defined prior to learning the Transition Rule.
Let $\mu$ denote the distribution of control single-cell images and $\nu_c$ the distribution under perturbation condition $c$. Their pushforwards under $\phi$ are
\begin{equation}
    \mu_z = \phi_\sharp \mu,
    \qquad
    \nu_{c,z} = \phi_\sharp \nu_c.
\end{equation}
To characterize the empirically supported cellular manifold, we construct a latent support bank
\begin{equation}
\mathcal Z_{\mathrm{emp}} = \{\phi(x_i)\}_{i=1}^N.
\end{equation}
Regions near $\mathcal Z_{\mathrm{emp}}$ correspond to morphologies observed in real data.

\subsection{Transition Rule: Schr\"odinger Bridge Transport}
Given the fixed geometry defined by the State Unit, the Transition Rule specifies stochastic transport between $\mu_z$ and $\nu_{c,z}$.
We model latent dynamics via the controlled stochastic differential equation:
\begin{equation}
dZ_t = u_t(Z_t)\,dt + \sigma\, dW_t,
\qquad
Z_0 \sim \mu_z.
\end{equation}
The objective is to identify a drift field $u_t$ such that:
\begin{equation}
Z_1 \sim \nu_{c,z}.
\end{equation}
Endpoint-only marginal matching is underdetermined before a reference process or path criterion is specified. A classical Schr\"odinger Bridge becomes well-defined once a Brownian reference process and boundary marginals are fixed:
\begin{equation}
\min_{\mathbb P^u}
\mathrm{KL}(\mathbb P^u \| \mathbb P^0)
\quad
\text{s.t. }
Z_0 \sim \mu_z,\;
Z_1 \sim \nu_{c,z},
\end{equation}
which is equivalent to the stochastic control objective~\cite{chen2016relation}
\begin{equation}
\min_{u}
\mathbb E
\left[
\frac{1}{2\sigma^2}
\int_0^1
\|u_t(Z_t)\|^2 dt
\right].
\end{equation}
FreeBridge follows this endpoint-constrained bridge view but augments it with a data-driven state cost that anchors intermediate paths to the empirical cellular morphology manifold, yielding a generalized Schr\"odinger Bridge formulation with task-specific state cost~\cite{liu2024generalized}.
The stochastic term $\sigma dW_t$ captures intrinsic cellular heterogeneity. Deterministic transport yields a single trajectory per initial state, whereas the stochastic formulation permits condition-consistent variability.

\subsection{Geometric Regularization of Intermediate Dynamics}
Once the boundary marginals are fixed, the intermediate marginals of the bridge are determined by the chosen reference process; departing from a purely diffusive reference therefore requires an explicit path criterion. In cellular modeling, intermediate states correspond to putative morphologies along the perturbation response and should remain consistent with empirically observed single-cell structures.  We therefore anchor stochastic transport to the geometry defined by the State Unit.  For any latent state $z$, define the empirical support distance:
\begin{equation}
d_{\mathrm{emp}}(z)
=
\min_{z_i \in \mathcal Z_{\mathrm{emp}}}
\|z - z_i\|.
\end{equation}
This nearest-neighbor statistic~\cite{biau2015lectures} serves as a nonparametric proxy for density support~\cite{silverman1986density}. Regions distant from $\mathcal Z_{\mathrm{emp}}$ correspond to latent configurations not supported by observed cells. We penalize deviations along the trajectory:
\begin{equation}
\mathcal L_{\mathrm{support}}
=
\mathbb E
\left[
\int_0^1
d_{\mathrm{emp}}(Z_t)^2 dt
\right].
\end{equation}
This regularization biases intermediate marginals toward empirically supported regions without imposing sample-level pairing constraints (Fig.~\ref{fig:intermediate_manifold}).

\myparagraph{Practical estimation.}
Direct evaluation of $d_{\mathrm{emp}}(z)$ against the full support bank requires $O(N)$ comparisons per query. During training, we approximate the population support using the union of control and perturbed embeddings in the current minibatch,
$\mathcal Z_B=\{z_0^{(i)}\}_{i=1}^B \cup \{z_1^{(i)}\}_{i=1}^B$.
For each intermediate state $Z_t^{(i)}$, we compute
\begin{equation}
\widehat d_{\mathrm{emp}}(Z_t^{(i)})
=
\min_{z_j \in \mathcal Z_B}
\|Z_t^{(i)}-z_j\|_2 .
\end{equation}
This yields a stochastic minibatch approximation to the population state-cost term under i.i.d. sampling. The pairwise distances are computed with batched matrix operations (\texttt{torch.cdist}), resulting in $O(B^2)$ batch complexity. The nearest-neighbor index is treated as non-differentiable, while gradients propagate through the selected squared distance.

\myparagraph{Interior Dynamics of Schr\"odinger Bridges.}
The Schr\"odinger Bridge constrains boundary marginals; given a fixed reference process, the interior marginals are then determined, so altering the empirical morphology of intermediate states requires an explicit state cost rather than endpoint supervision alone~\cite{chen2016relation,leonard2013survey}.  Endpoint alignment therefore does not, by itself, fix the desired transition geometry on the data manifold.  Figure~\ref{fig:intermediate_manifold} shows this effect on a synthetic manifold.  Even when boundary distributions coincide, transport under a diffusive reference may traverse sparsely supported regions of latent space.  Such trajectories preserve endpoint statistics yet alter interior structure. The support regularization restricts stochastic evolution to the empirical cellular manifold.

\subsection{FreeBridge Objective}
The FreeBridge objective is defined as a generalized Schr\"odinger Bridge with task-specific state cost $V_t(z)=\lambda\,d_{\mathrm{emp}}(z)^2$:
\begin{equation}
\min_{u} \;
\mathbb E \!\left[
\frac{1}{2\sigma^2} \int_0^1 \|u_t(Z_t)\|^2 dt
+ \lambda \int_0^1 d_{\mathrm{emp}}(Z_t)^2 dt
\right]
\;\; \text{s.t.} \;\;
Z_0 \sim \mu_z,\; Z_1 \sim \nu_{c,z}.
\end{equation}
\myparagraph{Intra-batch pairing for endpoint supervision.}
Endpoint populations are unpaired, but training requires batch-level tuples.
For each condition $c$, we sample $\{z_0^{(i)}\}_{i=1}^B \sim \mu_z$ and $\{z_1^{(i)}\}_{i=1}^B \sim \nu_{c,z}$ independently and pair them by index after random shuffling.
This stochastic coupling is used for optimization only and does not impose identity-level correspondence or constrain intermediate dynamics.

\subsection{Numerical Approximation and Optimization}
We parameterize the drift as $u_\theta(z,t,c)$ using a time-conditioned neural network. The continuous process is discretized via Euler--Maruyama:
\begin{equation}
Z_{t+\Delta t}
=
Z_t
+
u_\theta(Z_t,t,c)\Delta t
+
\sigma \sqrt{\Delta t}\,\epsilon,
\quad
\epsilon \sim \mathcal N(0,I).
\end{equation}
With $K$ Euler--Maruyama steps and $\Delta t = 1/K$, the training objective is approximated as
\begin{equation}
\widehat{\mathcal L}(\theta)
=
\frac{1}{B}\sum_{i=1}^B
\sum_{k=0}^{K-1}
\left[
\frac{\Delta t}{2\sigma^2}
\|u_\theta(Z_{t_k}^{(i)},t_k,c)\|_2^2
+
\lambda \Delta t\,
\widehat d_{\mathrm{emp}}(Z_{t_k}^{(i)})^2
\right]
+
\mathcal L_{\mathrm{end}},
\end{equation}
where the endpoint term $\mathcal L_{\mathrm{end}}$ aligns the terminal generated batch with samples from $\nu_{c,z}$ in the same latent metric used for evaluation. All endpoint samples are drawn independently; the random intra-batch pairing is used only as an optimization device and does not imply observed cell-level correspondences.

\begin{figure}[t]
\centering
\includegraphics[width=0.85\linewidth]{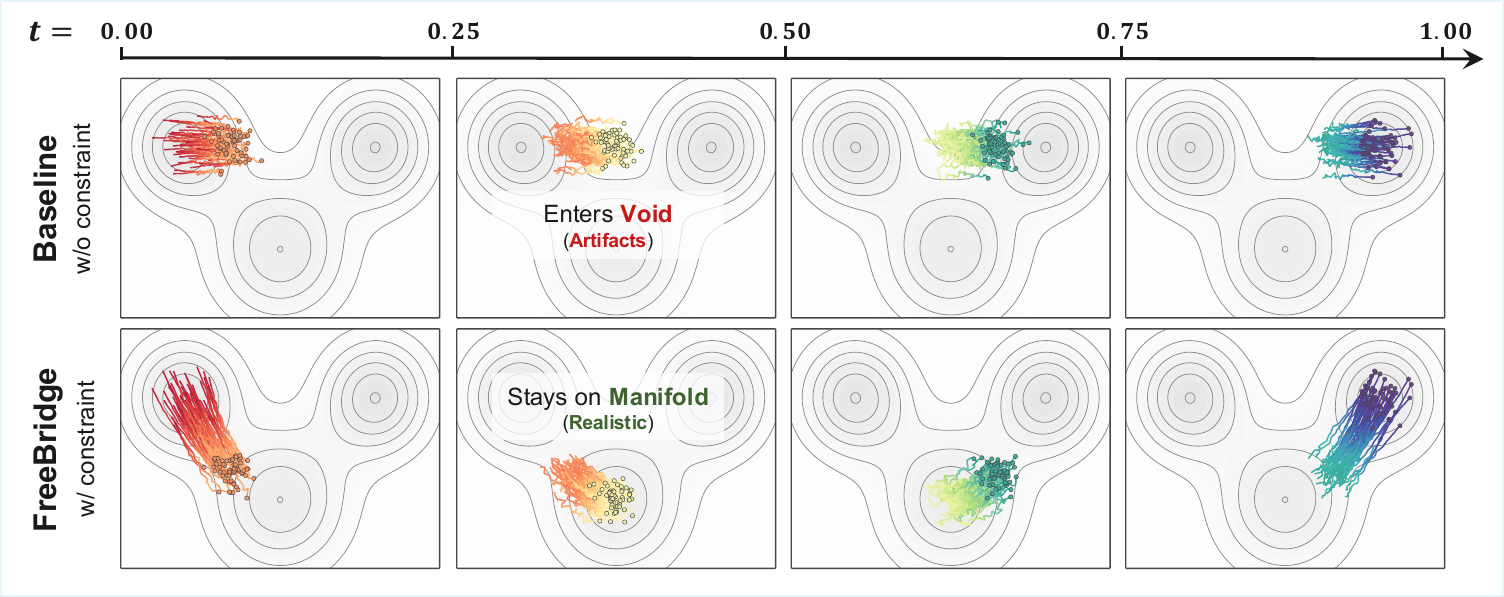}
\vspace{-10pt}
\caption{
\textbf{Intermediate trajectories under endpoint-constrained transport.}
Columns show increasing time along the stochastic evolution. Top (unconstrained): endpoints align, but intermediate states traverse low-support regions. Bottom (FreeBridge): support regularization prevents such excursions and preserves trajectory consistency.
}
\label{fig:intermediate_manifold}
\vspace{-5pt}
\end{figure}

\section{Experiments}
\label{sec:experiments}

\subsection{Experimental Setup}

\myparagraph{Datasets and Baselines.}
We evaluate FreeBridge on three widely used high-content imaging benchmarks: BBBC021~\cite{ljosa2012annotated}, RxRx1~\cite{sypetkowski2023rxrx1}, and JUMP~\cite{chandrasekaran2023jump}. These datasets cover chemical and genetic perturbations across multiple experimental conditions. We compare against recent morphology prediction methods, including PhenDiff~\cite{bourou2024phendiff}, IMPA~\cite{palma2025predicting}, and CellFlux~\cite{zhang2025cellflux}.

\myparagraph{Implementation Details.}
All methods are evaluated under a unified single-cell pipeline. Images are instance-segmented and cropped using the same preprocessing procedure. This ensures that differences in performance arise from transition modeling rather than data preparation.
The encoder $\phi$ is a frozen ResNet-50 backbone (latent dimension $d=2048$), and latent embeddings are precomputed for all segmented single-cell crops ($96\times96$). The drift network is a time-conditioned multilayer perceptron (hidden width $512$, $6$ layers) that takes $(z,t,c)$ as input. We optimize with Adam~\cite{kingma2014adam} using batch size $B=256$, learning rate $3\times10^{-4}$, and $K=50$ Euler--Maruyama discretization steps during training, with a minibatch support bank described in Sec. 2. The support weight $\lambda_{\mathrm{bank}}$, diffusion noise $\sigma$, and inference budget follow the default configuration; sensitivity to each is reported in Fig.~\ref{fig:param_analysis}. All reported metrics use 5k generated samples per condition and are averaged over three random seeds.
Endpoint fidelity is measured using Fr\'echet Inception Distance (FID)~\cite{heusel2017gans} and Kernel Inception Distance (KID)~\cite{binkowski2018demystifying}. To assess trajectory consistency, we report the Support Violation Rate ($\mathcal{R}_{\mathrm{viol}}$), defined as the proportion of intermediate states whose feature representations fall outside the empirical morphology support. Because fixed-cell assays do not provide observed single-cell trajectories, no metric can directly verify true temporal ordering at the individual-cell level; $\mathcal{R}_{\mathrm{viol}}$ is therefore a support-feasibility metric and should not be interpreted as direct evidence of biological temporal correctness. Semantic retention is evaluated using Mechanism-of-Action (MoA) classification accuracy obtained from a linear probe trained on real perturbed samples. Unless otherwise stated, official hyperparameters are used for baseline methods.

\begingroup
\setlength{\tabcolsep}{0pt}
\begin{table}[t]
\centering
\caption{\textbf{Quantitative comparison across datasets (5k generated samples per condition).}
We report overall and condition-level FID/KID together with the Support Violation Rate ($\mathcal{R}_{\mathrm{viol}}$).
All methods are evaluated under an identical single-cell preprocessing and inference protocol.
FreeBridge achieves improved endpoint fidelity while maintaining lower intermediate support violations.}
\label{tab:main_results}
\resizebox{\textwidth}{!}{%
\begin{tabular}{@{}l ccccc cccc cccc@{}}
\toprule
& \multicolumn{5}{c}{\textbf{BBBC021}} & \multicolumn{4}{c}{\textbf{RxRx1}} & \multicolumn{4}{c}{\textbf{JUMP}} \\
\cmidrule(lr){2-6}\cmidrule(lr){7-10}\cmidrule(lr){11-14}
\textbf{Method}
& \textbf{FID$_o \downarrow$} & \textbf{FID$_c \downarrow$} & \textbf{KID$_o \downarrow$} & \textbf{KID$_c \downarrow$} & $\boldsymbol{\mathcal{R}_{\mathrm{viol}} \downarrow}$
& \textbf{FID$_o \downarrow$} & \textbf{FID$_c \downarrow$} & \textbf{KID$_o \downarrow$} & \textbf{KID$_c \downarrow$}
& \textbf{FID$_o \downarrow$} & \textbf{FID$_c \downarrow$} & \textbf{KID$_o \downarrow$} & \textbf{KID$_c \downarrow$} \\
\midrule
PhenDiff & 48.25 & 107.50 & 3.08 & 3.26 & 0.38 & 67.12 & 175.62 & 5.28 & 5.34 & 48.57 & 126.52 & 5.13 & 5.23 \\
IMPA & 33.91 & 75.76 & 2.74 & 2.79 & 0.34 & 39.94 & 163.91 & 2.89 & 2.87 & 14.81 & 101.03 & 1.10 & 1.00 \\
CellFlux & 18.72 & 56.80 & 1.62 & 1.59 & 0.31 & 33.00 & 162.85 & 2.40 & 2.41 & 9.00 & 83.21 & 0.65 & 0.67 \\
\midrule
FreeBridge & \textbf{15.35} & \textbf{47.28} & \textbf{1.10} & \textbf{1.02} & \textbf{0.11}
& \textbf{28.94} & \textbf{147.33} & \textbf{1.98} & \textbf{2.03}
& \textbf{8.51} & \textbf{79.05} & \textbf{0.59} & \textbf{0.53} \\
\bottomrule
\end{tabular}}
\vspace{-5pt}
\end{table}
\endgroup

\subsection{Results}

\begin{figure*}[t]
\centering
\includegraphics[width=0.85\textwidth]{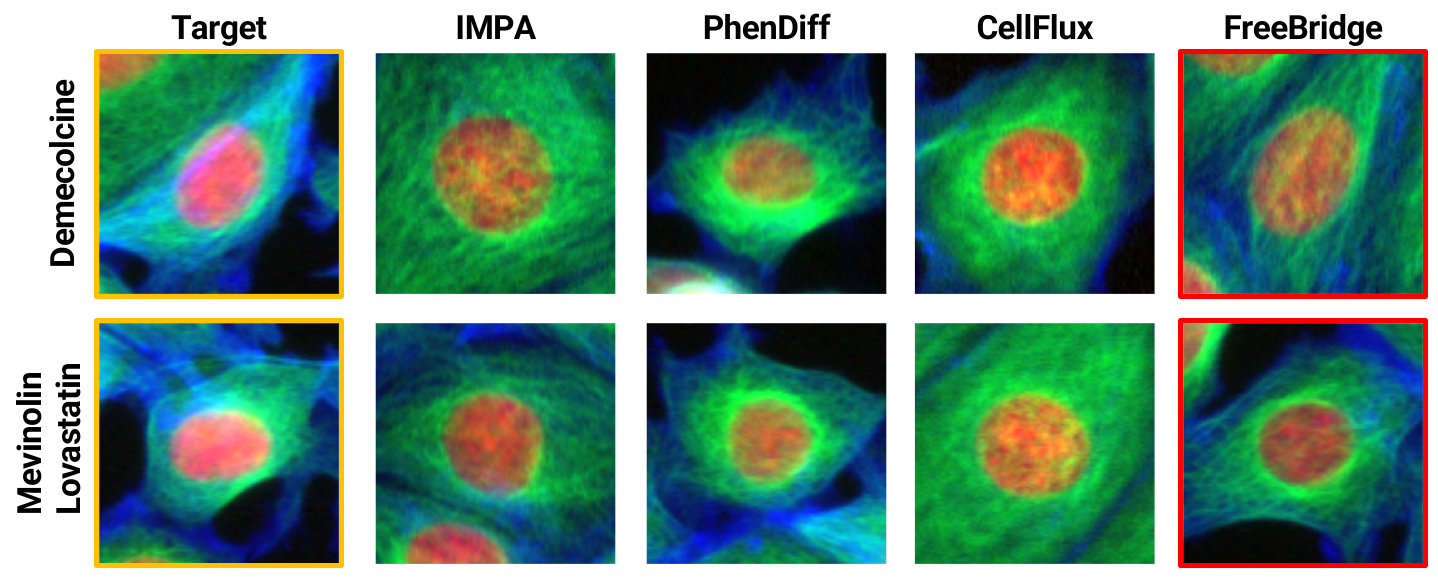}
\vspace{-8pt}
\caption{\textbf{Qualitative comparison of endpoint morphology on BBBC021.}
Representative single-cell crops for two perturbations are shown.
Columns display real targets and samples from IMPA, PhenDiff, CellFlux, and FreeBridge under the unified pipeline.
FreeBridge preserves compound-specific structures (e.g., nuclear compaction under Demecolcine and cytoplasmic shrinkage under Mevinolin), whereas baselines show texture smoothing or incomplete perturbation patterns.}
\label{fig:qual_bbbc021}
\vspace{-5pt}
\end{figure*}

\myparagraph{Quantitative and Qualitative Comparisons.}
Table~\ref{tab:main_results} reports quantitative results across datasets. Fig.~\ref{fig:overview}(b) provides a compact summary on BBBC021.
FreeBridge achieves strongest endpoint fidelity on all three benchmarks under the matched evaluation protocol. In addition to improvements in FID and KID, FreeBridge exhibits substantially lower $\mathcal{R}_{\mathrm{viol}}$ on BBBC021, indicating that intermediate states remain closer to empirically observed morphology regions.
Figure~\ref{fig:qual_bbbc021} shows representative examples on BBBC021.
The visual patterns align with quantitative results, with clearer preservation of compound-specific morphology.

\myparagraph{Intermediate Trajectory Consistency.}
As illustrated in Fig.~\ref{fig:intermediate_manifold}, endpoint alignment alone does not constrain interior geometry.
On real data, models without geometric anchoring may enter low-support regions of the single-cell manifold at intermediate times, producing distortions not captured by endpoint metrics.
Table~\ref{tab:feasibility} quantifies this effect: increasing $\lambda_{\mathrm{bank}}$ reduces both the feasibility energy $\mathcal{E}_{\mathrm{feas}}$ and the support violation rate $\mathcal{R}_{\mathrm{viol}}$, confirming that the support prior improves intermediate validity. As noted above, this measures support feasibility rather than verified temporal ordering.

\myparagraph{Semantic Retention.}
We further evaluate MoA classification accuracy to quantify preservation of condition-specific information. On BBBC021, removing geometric regularization reduces MoA accuracy from $72.08\%$ to $65.08\%$, indicating constraining intermediate evolution contributes to preserving mechanistic signal.

\begin{table}[t]
\centering
\caption{\textbf{Trajectory feasibility on BBBC021.}
All variants share identical architectures and differ only in the support weight $\lambda_{\mathrm{bank}}$.
Metrics are computed over $T=11$ time points per trajectory and reported as the median over three seeds.
Higher $\lambda_{\mathrm{bank}}$ improves geometric feasibility. Lower is better.}
\label{tab:feasibility}
\setlength{\tabcolsep}{0.0pt}
\resizebox{0.6\textwidth}{!}{
\begin{tabular}{l c c}
\toprule
\textbf{Variant} & $\mathcal{E}_{\mathrm{feas}}\downarrow$ & $\mathcal{R}_{\mathrm{viol}}\downarrow$ \\
\midrule
$\lambda_{\mathrm{bank}}=0$ (Unconstrained) & 2.48 & 0.31 \\
FreeBridge ($\lambda_{\mathrm{bank}}=0.3$) & 1.85 & 0.19 \\
\textbf{FreeBridge (Default, $\lambda_{\mathrm{bank}}=0.5$)} & \textbf{1.32} & \textbf{0.11} \\
\bottomrule
\end{tabular}}
\vspace{-5pt}
\end{table}

\subsection{Ablation}

\myparagraph{Effect of Regularizations.}
Table~\ref{tab:ablation} reports ablation results on BBBC021. Removing instance segmentation degrades both fidelity and MoA accuracy, suggesting that explicit single-cell state specification improves conditional modeling. Removing the support regularization increases $\mathcal{R}_{\mathrm{viol}}$ and reduces semantic retention. These observations are consistent with the role of explicit single-cell geometry and geometric regularization in constraining intermediate evolution.

\begingroup
\setlength{\tabcolsep}{0pt}
\begin{table}[t]
\centering
\caption{\textbf{Ablation study on BBBC021.}
We examine the effect of removing intra-batch pairing, support regularization,
and instance segmentation.
Both geometric anchoring and explicit single-cell state specification
contribute to improved fidelity and semantic retention.
Full-model results are from an independent ablation run and may differ
marginally from Table~\ref{tab:main_results}.}
\label{tab:ablation}
\resizebox{0.65\textwidth}{!}{
\begin{tabular}{l cccc}
\toprule
Variant & FID$_o \downarrow$ & FID$_c \downarrow$ & MoA $\uparrow$ & KID$_o \downarrow$ \\
\midrule
w/o intra-batch pairing & 19.94 & 60.81 & 63.27 & 1.81 \\
w/o support cost $V(\cdot)$ & 17.18 & 54.37 & 65.08 & 1.55 \\
w/o instance segmentation & 28.54 & 82.15 & 51.34 & 2.45 \\
\midrule
Full model & \textbf{15.37} & \textbf{47.41} & \textbf{72.08} & \textbf{1.13} \\
\bottomrule
\end{tabular}}
\end{table}
\endgroup

\begin{figure}[t]
\centering
\includegraphics[width=0.9\linewidth]{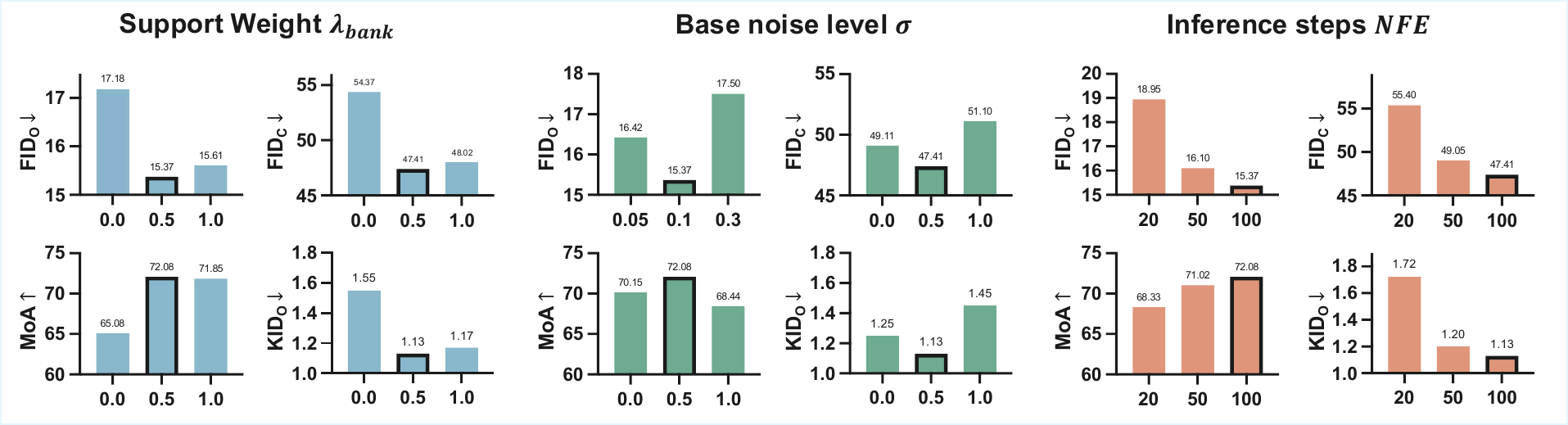}
\vspace{-5pt}
\caption{\textbf{Hyperparameter sensitivity on BBBC021.}
Performance is evaluated as a function of the support weight $\lambda_{\mathrm{bank}}$,
diffusion noise $\sigma$, and inference steps (NFE).
Results indicate stable endpoint fidelity across a range of settings,
with improved semantic retention at moderate regularization strength.}
\label{fig:param_analysis}
\vspace{-5pt}
\end{figure}

\myparagraph{Hyperparameter Sensitivity.}
Fig.~\ref{fig:param_analysis} shows sensitivity to the support weight $\lambda_{\mathrm{bank}}$, diffusion noise $\sigma$, and inference steps. Performance remains stable across a range of settings, with moderate $\lambda_{\mathrm{bank}}$ values providing improved MoA retention while maintaining endpoint fidelity.

\section{Conclusion}
We introduced FreeBridge, a Schr\"odinger Bridge formulation for single-cell transition modeling under endpoint-only supervision. By coupling entropy-regularized stochastic transport with explicit single-cell state geometry, FreeBridge links distribution-level alignment with trajectory-level consistency within empirically observed morphology space. Across multiple large-scale perturbation benchmarks, FreeBridge improves endpoint fidelity under a unified evaluation protocol; on BBBC021, it further reduces intermediate support violations. These findings suggest that modeling biological transitions requires not only boundary alignment but also explicit geometric grounding of intermediate evolution. Such a formulation offers a principled basis for virtual cell simulation and related transition modeling problems.

\myparagraph{Limitations and cost.}
FreeBridge relies on a frozen encoder to define the latent cellular geometry; morphological factors absent from encoder pretraining may therefore not be recoverable through transport learning alone. The support-distance computation uses batched pairwise distances and scales as $O(B^2)$ with minibatch size. This cost is manageable at the batch sizes used here, but larger support banks may require approximate nearest-neighbor search or memory-aware batching.

\begin{credits}
\subsubsection{\discintname}
The authors have no competing interests to declare that are relevant to the content of this article.
\end{credits}

\bibliographystyle{splncs04}
\bibliography{references}

\begin{thebibliography}{10}
\providecommand{\url}[1]{\texttt{#1}}
\providecommand{\urlprefix}{URL }
\providecommand{\doi}[1]{https://doi.org/#1}

\bibitem{biau2015lectures}
Biau, G., Devroye, L.: Lectures on the nearest neighbor method, vol.~246.
  Springer (2015)

\bibitem{binkowski2018demystifying}
Bi{\'n}kowski, M., Sutherland, D.J., Arbel, M., Gretton, A.: Demystifying mmd
  gans. arXiv preprint arXiv:1801.01401  (2018)

\bibitem{bourou2024phendiff}
Bourou, A., Boyer, T., Gheisari, M., Daupin, K., Dubreuil, V., De~Thonel, A.,
  Mezger, V., Genovesio, A.: Phendiff: Revealing subtle phenotypes with
  diffusion models in real images. In: International Conference on Medical
  Image Computing and Computer-Assisted Intervention. pp. 358--367. Springer
  (2024)

\bibitem{bunne2024build}
Bunne, C., Roohani, Y., Rosen, Y., Gupta, A., Zhang, X., Roed, M., Alexandrov,
  T., AlQuraishi, M., Brennan, P., Burkhardt, D.B., et~al.: How to build the
  virtual cell with artificial intelligence: Priorities and opportunities. Cell
   \textbf{187}(25),  7045--7063 (2024)

\bibitem{carpenter2007image}
Carpenter, A.E.: Image-based chemical screening. Nature Chemical Biology
  \textbf{3}(8),  461--465 (2007)

\bibitem{chandrasekaran2023jump}
Chandrasekaran, S.N., Ackerman, J., Alix, E., Ando, D.M., Arevalo, J., Bennion,
  M., Boisseau, N., Borowa, A., Boyd, J.D., Brino, L., et~al.: Jump cell
  painting dataset: morphological impact of 136,000 chemical and genetic
  perturbations. BioRxiv pp. 2023--03 (2023)

\bibitem{chen2016relation}
Chen, Y., Georgiou, T.T., Pavon, M.: On the relation between optimal transport
  and schr{\"o}dinger bridges: A stochastic control viewpoint. Journal of
  Optimization Theory and Applications  \textbf{169}(2),  671--691 (2016)

\bibitem{heusel2017gans}
Heusel, M., Ramsauer, H., Unterthiner, T., Nessler, B., Hochreiter, S.: Gans
  trained by a two time-scale update rule converge to a local nash equilibrium.
  Advances in neural information processing systems  \textbf{30} (2017)

\bibitem{ho2020denoising}
Ho, J., Jain, A., Abbeel, P.: Denoising diffusion probabilistic models.
  Advances in neural information processing systems  \textbf{33},  6840--6851
  (2020)

\bibitem{kingma2014adam}
Kingma, D.P., Ba, J.: Adam: A method for stochastic optimization. arXiv
  preprint arXiv:1412.6980  (2014)

\bibitem{leonard2013survey}
L{\'e}onard, C.: A survey of the schr{\"o}dinger problem and some of its
  connections with optimal transport. arXiv preprint arXiv:1308.0215  (2013)

\bibitem{lipman2022flow}
Lipman, Y., Chen, R.T., Ben-Hamu, H., Nickel, M., Le, M.: Flow matching for
  generative modeling. arXiv preprint arXiv:2210.02747  (2022)

\bibitem{liu2024generalized}
Liu, G.H., Lipman, Y., Nickel, M., Karrer, B., Theodorou, E., Chen, R.T.:
  Generalized schr{\"o}dinger bridge matching. In: International Conference on
  Learning Representations. vol.~2024, pp. 14527--14552 (2024)

\bibitem{liu2022flow}
Liu, X., Gong, C., Liu, Q.: Flow straight and fast: Learning to generate and
  transfer data with rectified flow. arXiv preprint arXiv:2209.03003  (2022)

\bibitem{ljosa2012annotated}
Ljosa, V., Sokolnicki, K.L., Carpenter, A.E.: Annotated high-throughput
  microscopy image sets for validation. Nature methods  \textbf{9}(7), ~637
  (2012)

\bibitem{palma2025predicting}
Palma, A., Theis, F.J., Lotfollahi, M.: Predicting cell morphological responses
  to perturbations using generative modeling. Nature Communications
  \textbf{16}(1), ~505 (2025)

\bibitem{ren2025scale}
Ren, Q., Wang, Y., Guo, L., Zhang, W., Fan, Z., You, C.: Scale where it
  matters: Training-free localized scaling for diffusion models. arXiv preprint
  arXiv:2511.19917  (2025)

\bibitem{silverman1986density}
Silverman, B.W.: Density estimation for statistics and data analysis. Routledge
  (2018)

\bibitem{slepchenko2003quantitative}
Slepchenko, B.M., Schaff, J.C., Macara, I., Loew, L.M.: Quantitative cell
  biology with the virtual cell. Trends in cell biology  \textbf{13}(11),
  570--576 (2003)

\bibitem{song2020score}
Song, Y., Sohl-Dickstein, J., Kingma, D.P., Kumar, A., Ermon, S., Poole, B.:
  Score-based generative modeling through stochastic differential equations.
  arXiv preprint arXiv:2011.13456  (2020)

\bibitem{stringer2021cellpose}
Stringer, C., Wang, T., Michaelos, M., Pachitariu, M.: Cellpose: a generalist
  algorithm for cellular segmentation. Nature methods  \textbf{18}(1),
  100--106 (2021)

\bibitem{sun2025ouroboros}
Sun, S., Wang, Y., Zhang, H., Xiong, Y., Ren, Q., Fang, R., Xie, X., You, C.:
  Ouroboros: Single-step diffusion models for cycle-consistent forward and
  inverse rendering. In: Proceedings of the IEEE/CVF International Conference
  on Computer Vision (2025)

\bibitem{sypetkowski2023rxrx1}
Sypetkowski, M., Rezanejad, M., Saberian, S., Kraus, O., Urbanik, J., Taylor,
  J., Mabey, B., Victors, M., Yosinski, J., Sereshkeh, A.R., et~al.: Rxrx1: A
  dataset for evaluating experimental batch correction methods. In: Proceedings
  of the IEEE/CVF conference on computer vision and pattern recognition (2023)

\bibitem{wang2026let}
Wang, Y., Ma, Y., Li, W., You, C.: Let eeg models learn eeg. In: International
  conference on machine learning (2026)

\bibitem{zhang2025cellflux}
Zhang, Y., Su, Y., Wang, C., Li, T., Wefers, Z., Nirschl, J., Burgess, J.,
  Ding, D., Lozano, A., Lundberg, E., et~al.: Cellflux: Simulating cellular
  morphology changes via flow matching. arXiv preprint arXiv:2502.09775  (2025)

\end{thebibliography}

\end{document}